\documentclass[sigconf,natbib=false]{acmart}
\AtBeginDocument{%
  }

\setcopyright{acmlicensed}

\copyrightyear{2026}
\acmYear{2026}
\setcopyright{cc}
\setcctype{by}
\acmConference[SAC '26]{The 41st ACM/SIGAPP Symposium on Applied Computing}{March 23--27, 2026}{Thessaloniki, Greece}
\acmBooktitle{The 41st ACM/SIGAPP Symposium on Applied Computing (SAC '26), March 23--27, 2026, Thessaloniki, Greece}
\acmPrice{}
\acmDOI{10.1145/3748522.3779879}
\acmISBN{979-8-4007-2294-3/2026/03}

\RequirePackage[
  datamodel=acmdatamodel,
  style=acmnumeric,
  ]{biblatex}

\begin{document}

\title{Evaluating Austrian A-Level German Essays with Large Language Models for Automated Essay Scoring}
\renewcommand{\shorttitle}{AES for Austrian A-Level exams}

\author{Jonas Kubesch}
\email{jonas.kubesch@fh-salzburg.ac.at}
\orcid{0009-0009-7619-6592}
\affiliation{%
  \institution{Salzburg University of Applied Sciences}
  \city{Salzburg}
  \country{Austria}
}

\author{Lena Huber}
\email{lenamaria.huber@fh-salzburg.ac.at}
\orcid{0009-0006-0908-3975}
\affiliation{%
  \institution{Salzburg University of Applied Sciences}
  \city{Salzburg}
  \country{Austria}
}
\author{Clemens Havas}
\email{clemens.havas@fh-salzburg.ac.at}
\orcid{0000-0003-0390-5094}
\affiliation{%
  \institution{Salzburg University of Applied Sciences}
  \city{Salzburg}
  \country{Austria}
}
\renewcommand{\shortauthors}{J. Kubesch et al.}

\begin{abstract}
Automated Essay Scoring (AES) has been explored for decades with the goal to support teachers by reducing grading workload and mitigating subjective biases. While early systems relied on handcrafted features and statistical models, recent advances in Large Language Models (LLMs)
have made it possible to evaluate student writing with unprecedented flexibility. This paper investigates the application of state-of-the-art open-weight LLMs for the grading of Austrian A-level German texts, with a particular focus on rubric-based evaluation.
A dataset of 101 anonymised student exams across three text types was processed and evaluated. Four LLMs, DeepSeek-R1 32b, Qwen3 30b, Mixtral 8x7b and LLama3.3 70b, were evaluated with different contexts and prompting strategies.
The LLMs were able to reach a maximum of 40.6\% agreement with the human rater in the rubric-provided sub-dimensions, and only 32.8\% of final grades matched the ones given by a human expert.
The results indicate that even though smaller models are able to use standardised rubrics for German essay grading, they are not accurate enough to be used in a real-world grading environment. 
\end{abstract}



\maketitle
\section{Introduction}
\label{section:Introduction}
Writing essays is an essential part of modern education, as they often appear in class assignments, homework, and exams. For students, writing texts by hand is a time-consuming task and tests their skills in various areas, such as grammar, reading comprehension, concise writing, and text structure. This also applies to essay graders, who spend many hours reviewing a single set of assignments for one class. According to \textcite{mussmann2016niedersachsische}, German high school teachers spend an average of 13.57\% of their work time on grading and providing feedback. By utilizing novel techniques in the field of AES, the time spent for grading could be shortened drastically, allowing teachers to direct more focus to planning in-class activities and reducing overall workload. 

Applying LLMs for AES tasks has been done before \autocite{stahl2024exploring, pack2024large}, with use cases ranging from providing a singular grade, over using a predefined grading rubric to a deep explanation of different aspects, strengths and weaknesses of the candidate text through written feedback. Compared to previous work on essay grading, this paper addresses a wider variety of text types that can be assigned to A-level students, each following its own set of grading criteria. Therefore, the AES system must be capable of identifying, recalling, and correctly applying the appropriate rubrics and presets to all seven A-level types of texts.

For this paper, a comparison of zero-shot and few-shot prompting, with and without Chain-of-Thought (CoT) reasoning, is being conducted. For document retrieval, a Retrieval Augmented Generation (RAG)-based approach, retrieving similar documents to the candidate text, is compared to a fixed-context, with texts selected to depict the whole grading spectrum.  

\section{Related Work}
\label{section:RelatedWorks}
\subsection{Pre-Transformer AES}
The possibility of computer-based AES was first proposed in Project Essay Grading \autocite{page1966imminence}, which showed that computers can produce results indistinguishable from those of expert human graders.
Over the years, many improvements have been made, such as latent semantic analysis (LSA) \autocite{deerwester1990indexing}, e-rater \autocite{burstein2003rater}
and IntelliMetric \autocite{elliot2003intellimetric}.
With the introduction of word embedding by \textcite{mikolov2013efficient}, AES tasks were able to utilise more complex Neural Network-based models. Convolutional Neural Network (CNN) approaches \autocite{dong2016automatic} and its Recurrent Neural Network (RNN) pendants \autocite{taghipour2016neural}, as well as mixed method approaches \autocite{dong2017attention}, improved the quality of AES. However, they are limited by the ability to capture long-term relations between words (CNN) as well as inability to be parallelized (RNN).

\subsection{Transformers and Large Language Models}
Through the invention of the transformer architecture, \textcite{vaswani2017attention} ushered in a new era for AES, first, through the invention of the BERT architecture \autocite{devlin2019bert} and ultimately by the now omnipresent LLMs. Both architectures avoid the inherent restrictions of CNN and RNN models, allowing for large amounts of data to be processed in a relatively short amount of time. Early examples of LLMs that allow in-context learning are GPT-3 \autocite{brown2020language} and LLama \autocite{touvron2023llama}, which no longer require fine-tuning but instead leverage in-context learning and advanced prompting strategies. This paradigm shift marks the transition from task-specific pre-trained models to versatile LLMs, enabling zero-shot and few-shot AES with competitive accuracy and improved feedback generation.

Building on this advancement, \textcite[]{xiao2024automation} used a fine-tuned LLM as a co-grader to increase the performance of novice graders and help provide valuable feedback in less time. With sufficient training, they achieved scores close to those of state-of-the-art (SOTA) BERT models. \textcite{lee2024applying} demonstrate that incorporating CoT reasoning into LLM workflows increases scoring precision when combined with context relevant to the task. They also compared zero-shot and few-shot variants, deeming few-shot prompting with CoT and relevant context to be the best. Furthermore, \textcite{mizumoto2023exploring} ran 12,100 samples of the TOEFL exam against the GPT-3.5-based text-davinci-003 model. Their findings include a strong reproducibility of evaluation results, with 100\% overlap in a second rerun within a 2-point difference on a 10-point scale. 

\subsection{AES for tasks in German}
Taking a closer look at research on German-speaking students, 
\textcite{ludwig2021automated} were among the first to test attention-based models for German essay grading tasks. They compared a regression-based Bag of Words model with two transformer models, bert-base-german-cased and gbert-base, on the task of classifying whether an email written by a student was formulated in a polite or impolite tone. The results heavily favoured the transformer models.

For automated essay grading to be applicable in A-level exams, it is also important to analyse the structure of longer texts. Different types of texts demand different formatting, which is why \textcite{stahl2024school} gathered 1,320 student essays with annotations for argumentative structure and overall text quality. They perform argument mining on their corpora alongside assigning an overall score using mDeBERTaV3 \autocite{he2021debertav3}.
Combining individual essay dimensions, such as arguments and discourse functions, with the mDeBERTaV3 model for essay scoring, yields better results than just using the essay scoring model without the additional information.

\textcite{schaller2024fairness} conducted an experiment about fairness in AES for German learners. They compared an SVM, a BERT and an LLM-based grading approach, investigating their fairness regarding students with different cognitive capabilities. Their findings conclude that each of the three models produced fair results, but only if the target group was reasonably represented in the training data. Otherwise, scoring accuracy diminished and fair grades could not be guaranteed. 

\textcite{sessler2025can} highlighted the issue of possible grader-related preferences in datasets commonly used for AES benchmarks, such as the ASAP dataset which only used two graders for their gold standard assessments \autocite{hamner2012asap}. They gathered 37 teachers to grade 20 texts and compared their results with the grades given by five LLMs. The three  OpenAI-based commercial LLMs outperformed LLama 3-70B and Mixtral 8x7B. However, \textcite{sessler2025can} did not employ any prompt engineering or CoT methodology and also abstained from few-shot prompting. Their study provides a baseline without prompt engineering, which this paper extends by testing CoT and few-shot prompting.

\textcite{firoozi2025using} investigate the issue of grading essays in multilingual countries such as Belgium. Therefore, they compare a BERT model trained on 104 languages called mBert to the language-agnostic BERT sentence embedding model LaBSE \autocite{feng2020language}. In their comparison, they found substantial overlap for both models with human graders for essays written in German, Italian and Czech. LaBSE scored slightly better than mBERT, indicating the usefulness of training BERT-based models to be language-agnostic for non-English tasks.

Taken together, these studies show that while German AES has advanced rapidly with transformers and LLMs, there remains a gap in applying and rigorously evaluating such models for A-level examinations, particularly under rubric-based conditions. This paper addresses this gap by comparing four open source state-of-the-art LLMs on the task of scoring texts written by Austrian students with regard to a predetermined, nationwide used, grading rubric.

\section{Datasets and Methods}
\label{section:Data_Methods}
The aim of this paper is to evaluate the usefulness of SOTA LLMs in grading Austrian A-level exams in the subject "German". First, an overview of the exam is given to introduce the reader to the modalities. Then, the possible types of text are introduced, with an emphasis on the types used in the experiment. Furthermore, a data description is given before the experiment setup is described.

\subsection{Introduction to Austrian A-levels}
According to the Austrian Federal Ministry of Education, Science and Research \cite{bmbwf2020beurteilung} there are currently seven types of text that may be used in A-level exams:

\begin{itemize}
    \item Erörterung (Discussion Essay)
    \item Kommentar (Commentary)
    \item Leserbrief (Letter to the Editor)
    \item Meinungsrede (Opinion Speech)
    \item Textanalyse (Text Analysis)
    \item Textinterpretation (Literary Interpretation)
    \item Zusammenfassung (Summary)
\end{itemize}

From these texts, three exercise packs, consisting of two text types with assignments, are formed. The A-level students need to choose one of the packs for their exam and fulfil the assignments.

The dataset provided by the team of the Standardised School-Leaving and Diploma Examination (SRDP), the committee for standardised A-level exams, contains solutions for the main exam dates of 2023 (Literary Interpretation, Letter to the Editor) \autocite{bmbwf2023angaben} and 2024 (Literary Interpretation, Commentary) \autocite{bmbwf2024angaben}. Below is a brief introduction to the text types

\begin{itemize}
    \item \textbf{Commentary:} A journalistic text type whose goal is to influence the reader's opinion. The author's personal stance towards the topic is being depicted in the text. The text should be suited for a public presentation, e.g. in a newspaper.
    \item \textbf{Letter to the Editor:} Presents the author's personal opinion towards a topic or event that is either commonly known or based upon a preceding article in the same medium, the letter will be published in. 
    \item \textbf{Literary Interpretation:} Describes its underlying text based on its literary features. The interaction between formal, linguistic and content-related aspects is uncovered to create a deeper understanding of the reference text. 
\end{itemize}

\subsection{Grading Austria A-levels}
As the exams are taken nation wide, guaranteeing comparability is a difficult task. Hence, in 2015, the "Zentralmatura", a centralised and standardised version of the A-levels, was rolled out in Austria. The topics and tasks are distributed by the SRDP committee, alongside one standardised grading rubric per task.

As usual in Austrian grading, a five-part grading scale is used, with 1 being the best and 5 the worst possible grade. For each of the four sections, at least a 4 has to be scored, otherwise the overall exam is assessed negatively. 
For each grading criteria, detailed descriptions exist for grades 1 to 4, while grade 5 is seen as a failure to fulfil one or more required structural or content-related requirements. 

The rubric is divided into two sections per text, resulting in four sections per exam.
In the main sections, K1 for the first text, K2 for the second text, two major sections, Content and Structure, are scored.
The remaining sections K3/1 and K3/2 are scoring Style and Expression, as well as Language Conventions for Task 1 and Task 2 respectively. In these sections, it is possible for either to be negative. However, if the other section is overwhelmingly positive, it is possible to sum up K3/1 and K3/2 to receive an overall positive score for the combined section K3.

Given that all sections, K1, K2 and K3, are scored positively, the average grade achieved across these sections results in the final exam grade.

\subsection{Dataset and LLMs}
The SRDP team provided 173 datasets, which were printed, redacted, and then re-scanned for data protection reasons, eliminating the possibility for standard PDF-text extraction. 28 sets contained incomplete grading data, and out of the remaining 145 texts, 44 were written by hand, often times in cursive. This, alongside the low quality of the provided scans, lead to the authors being unable to extract handwritten information without including major artifacts. This leaves 101 texts suitable for automated grading.

For technique evaluation, four LLMs are inspected: LLama3.3 70b, DeepSeek-R1 32b, Qwen3 30b and Mixtral 8x7b. The models were chosen based on their size and capability of understanding German texts. Only LLama3.3 was evaluated more thoroughly, as the other models showed significant flaws as discussed in section \ref{section:Discussion}.

\section{Implementation and Experiments}
\label{section:Implementation}
\subsection{Baseline}

\begin{figure}
    \centering
    \includegraphics[width=0.95\linewidth]{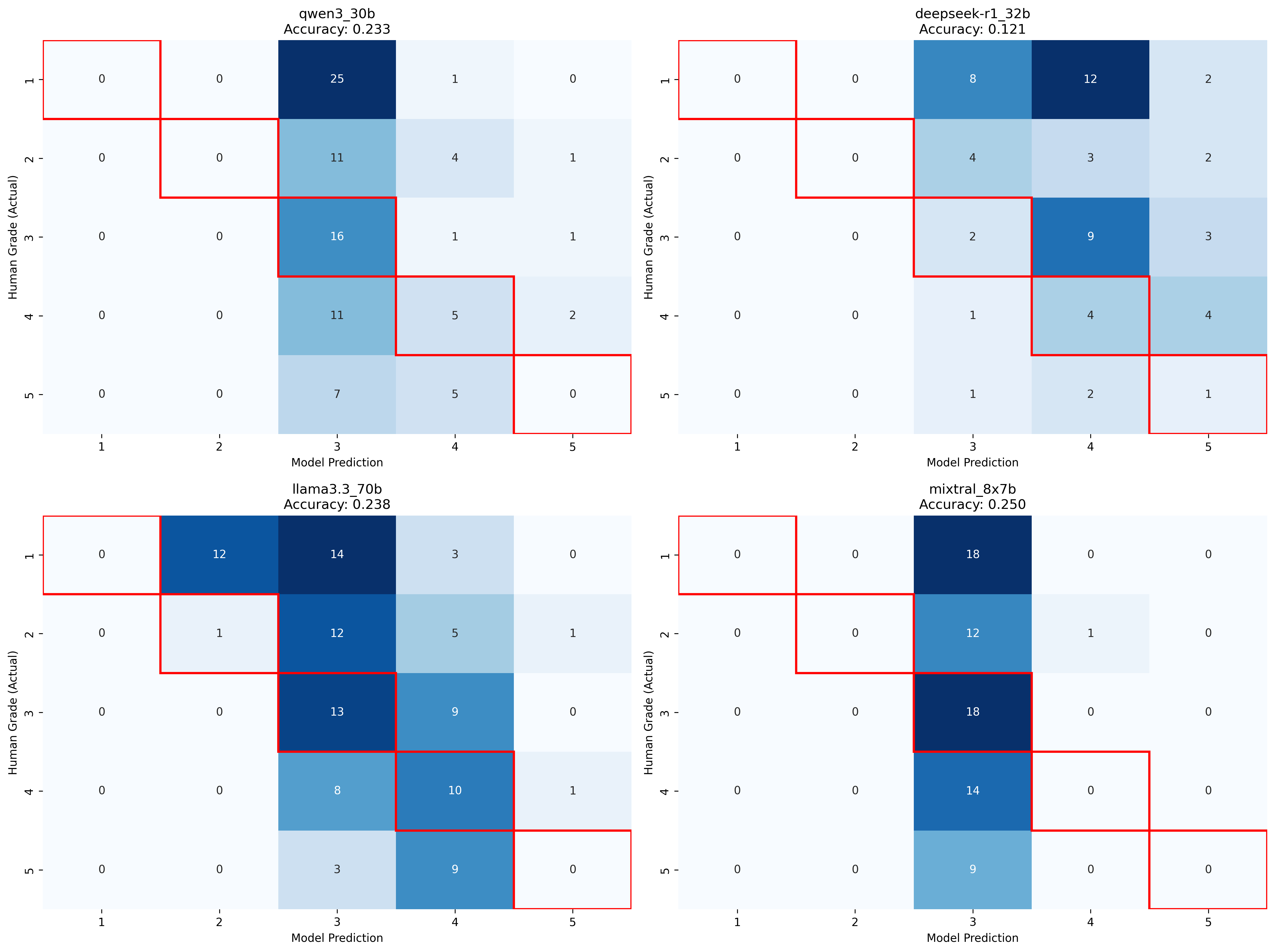}
    \caption{Confusion matrix for mismatches between human and model grades in the baseline setup.}
    \label{fig:v3_2_confusion}
\end{figure}

In a baseline attempt, a system prompt, containing the grading criteria and a desired JSON output schema, are implemented. The OCR-extracted student texts are passed to the LLM for grading, alongside a brief summary of the text type and information from the rubric, which indicates the criteria to be met for achieving a particular score. The difference between human and LLM grade can be seen in Figure \ref{fig:v3_2_confusion}.

\begin{figure}
    \centering
    \includegraphics[width=0.95\linewidth]{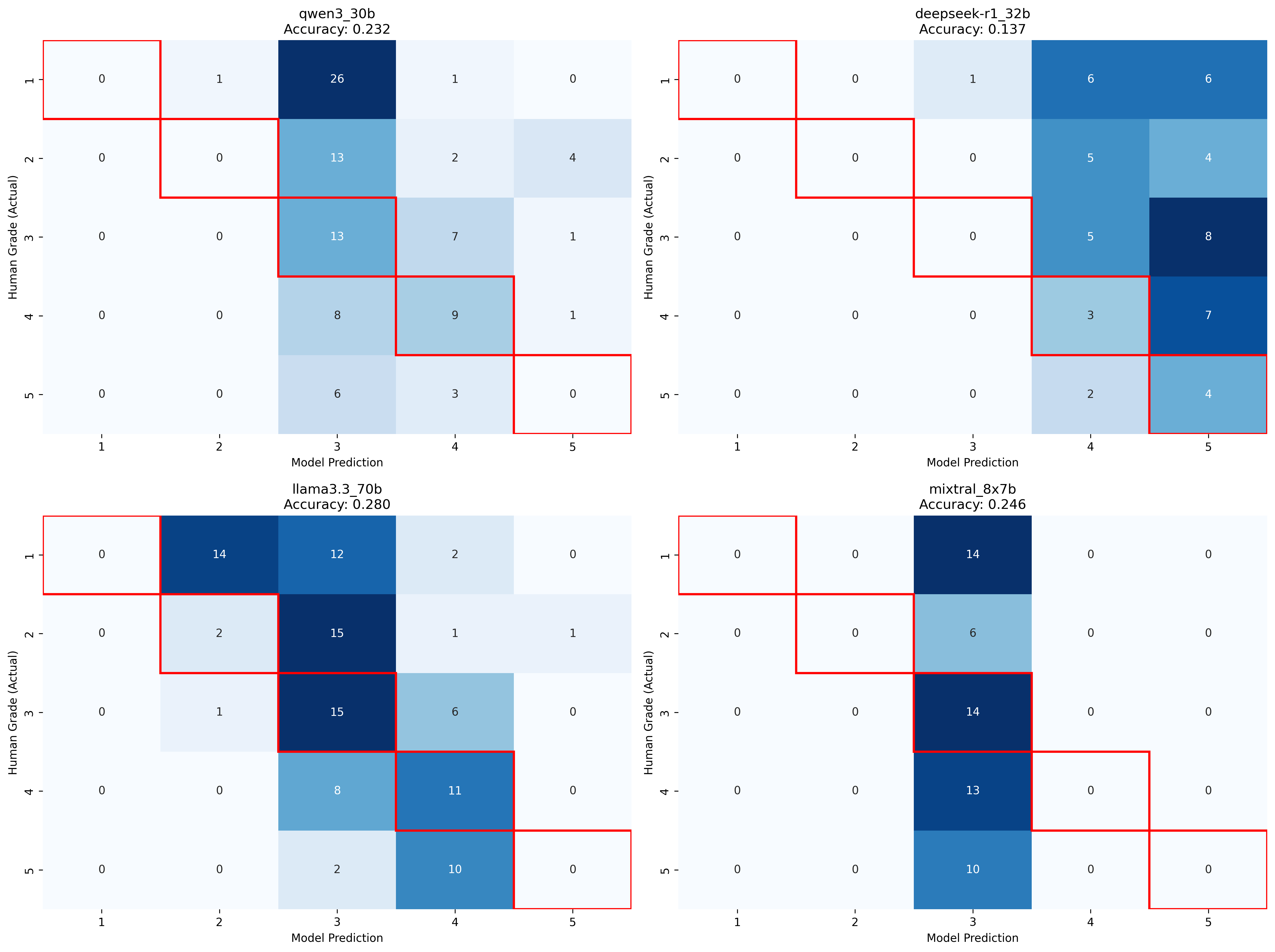}
    \caption{Confusion matrix for mismatches using RAG-Best-Average-Worst}
    \label{fig:v5-baw-confusion}
\end{figure}
\subsection{RAG for context}
So far, the LLMs have never seen the correct solution for a given task, which makes them incapable of knowing whether their applied grading standards are in line with those of human graders.

By incorporating a RAG module into the candidate LLMs, the majority of samples can be used for testing, compared to fine-tuning, where many are lost to the training process. Furthermore, a well-set-up RAG pipeline can handle data it has never seen before, be applied to different types and versions of LLMs, and incorporate changing types of text or grading criteria easily. These traits make it the superior choice over fine-tuning for the task at hand.

\begin{figure}
    \centering
    \includegraphics[width=1\linewidth]{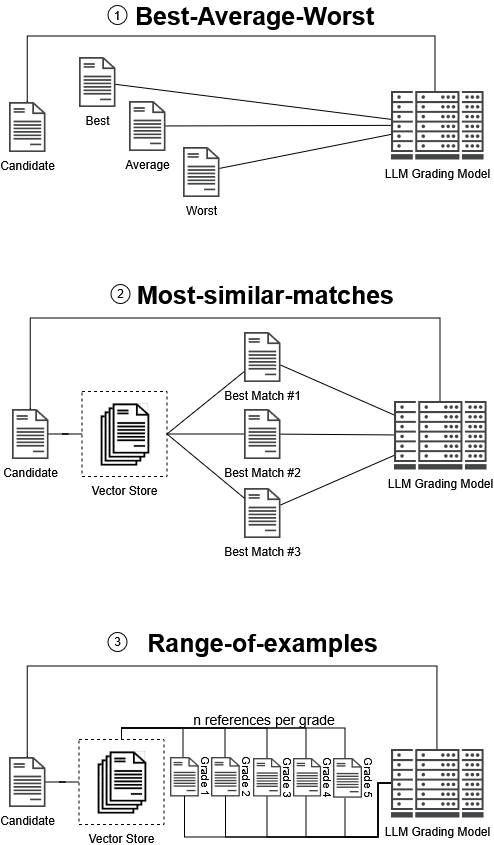}
    \caption{Visualization of context modes.}
    \label{fig:RAG-visualization}
\end{figure}

To investigate the efficiency of context passing, three experiment setups are tested, which are visualized in Figure \ref{fig:RAG-visualization}:
\begin{itemize}
    \item \textbf{Top: Best-Average-Worst} For each of the four year-task combinations, the best and worst graded texts, as well as the one scoring the most average grades across all grading dimensions, are passed as fixed contexts for every task. These are always the same for every sample text.

    \item \textbf{Middle: Most-similar-matches} Using the thought process of similar texts leading to similar grading, a vector store is set up with \textit{T-Systems-onsite/german-roberta-sentence-transformer-v2} encoding, and the most similar texts to the candidate exam are passed to the LLM. These texts change for every candidate text.
    \item \textbf{Bottom: Range-of-examples} As can be seen in Figure \ref{fig:v3_2_confusion} “extreme” grades, 1 and 5 are being severely under-represented in the grading process. The "Range-of-examples" approach hands sets of 5 texts, graded 1-5, to the models. Receiving every possible grade should encourage the LLMs to allow for more diverse grading. As in the Most-similar-matches approach, the best matching texts are chosen, up until a point where every grade is represented n-times, depending on the desired setting.
\end{itemize}

\subsection{Few-shot prompting}
\begin{figure}
    \centering
    \includegraphics[width=1\linewidth]{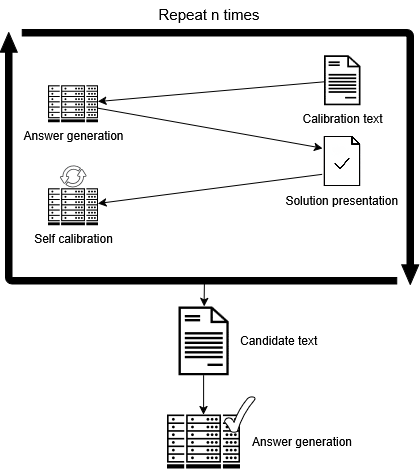}
    \caption{Workflow of few-shot in-context learning}
    \label{fig:Few-shot}
\end{figure}

RAG has increased the LLM's grading capabilities, however they still have problems with providing accurate results for perfect or failed exams. The needed context is provided, but gets lost due to excessive message lengths and no way for the LLMs to adjust their grading based on feedback. With turn based few-shot prompting, as can be seen in Figure \ref{fig:Few-shot}, the models can iteratively estimate the rating for several texts, and are provided with the correct solution after their guess.

This form of self-adjustment was tested with various combinations of accompanying prompts. Notably, adding context retrieved based on similarity to the candidate text did not provide good results. Neither Most-similar-matches nor Range-of-examples retrieval were beneficial to the calibration process. An initially acceptable baseline was frequently misled by either very positive or very negative examples, skewing the results towards the latest reference texts.

The Best-Average-Worst setup broke this pattern, by offering a solid, non-changing grading range, with a perfect, an average, and a failed text. Importantly, all four grading dimensions per text were marked as 1, 3 and 5 respectively, serving as a guideline.

Further experiments were conducted with texts which had all dimensions marked as 2 and 4 respectively, called 
All-Grade runs. The texts were presented from best to worst (1 to 5), resulting in a distribution skewed toward lower grades. To combat this, an inverted scheme (5 to 1) and mixed schemes (1,5,2,4,3) and (1,5,3), were also tested. The inverted scheme yielded results opposite to the base scheme, overrating texts, while the mixed schemes performed worse for good and bad texts.

In general, the base implementation of starting with good texts and iterating to worse texts showed the most stable performance in test runs, and was chosen for comparison on the complete dataset. 

Another discovery was made while inspecting the results of Best-Average-Worst versus All-Grade runs. The number of context texts provided differs: 3 versus 5. When inspecting the individual dimensions, Task 1, containing the longer text type, performed significantly worse when receiving All-Grade context, compared to Best-Average-Worst context. The inverse is true for Task 2. 

Consequently, a mixed method run has been conducted, using Best-Average-Worst for Task 1 and All-Grades for Task 2.

As a final technique, CoT is being applied to few-shot prompting. Preliminary testing for a test dataset showed no major improvements by adding CoT instructions, hence only one trial run was conducted with few-shot Best-Average-Worst prompting.

\section{Evaluation}
\label{section:Evaluation}
\subsection{Results}
For this paper, four different experimental setups, with 14 full-dataset results, have been created. For evaluation criteria, Quadratic Weighted Kappa (QWK, Table \ref{tab:qwk-compressed}), Mean Average Error (MAE), Pearson Correlation Coefficient (PCC) and accuracy (Table \ref{tab:acc-compressed}) have been chosen.

QWK measures inter rater agreement between the LLM output and the human gold standard, while MAE indicates the average error made by the model. PCC depicts the linear relation between predicted and gold standard grades. The accuracy metric measures the amount of correctly assigned grades, without taking into consideration the magnitude of error for misclassification.

Out of the four proposed LLMs, only LLama3.3 70b was able to perform the task of grading A-level essays in a stable and diverse manner. 

\begin{figure}
    \centering
    \includegraphics[width=0.9\linewidth]{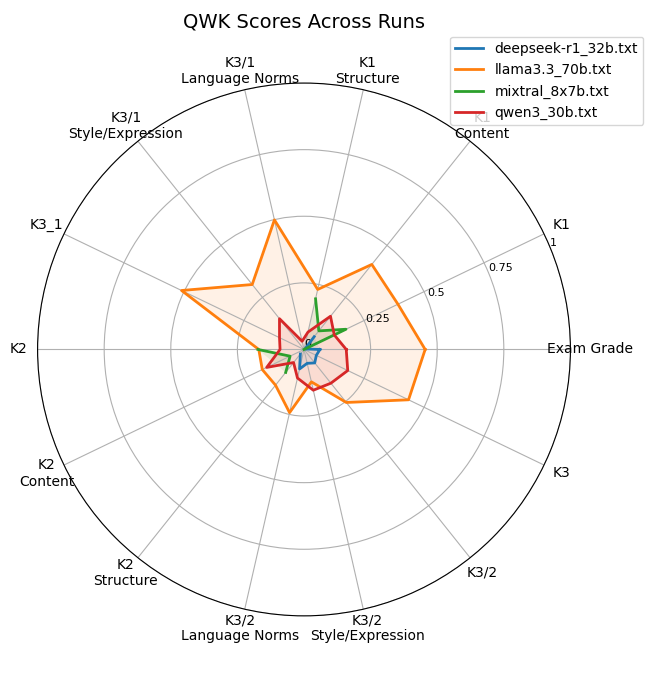}
    \caption{Comparison of QWK-scores for Best-Average-Worst context files with different LLMs}
    \label{fig:radar-baw}
\end{figure}

Running a preliminary test with a RAG Best-Average-Worst setup, the results in Figure \ref{fig:v5-baw-confusion} were achieved. As can be seen, compared to Figure \ref{fig:v3_2_confusion}, the results for LLama and Qwen rarely differ. The evaluation of Mixtral always resulted in the grade 3, and DeepSeek was also not able to provide a grade better than 3. Qwen was a bit more diverse and scored an accuracy of 0.232, but fell short to LLama3.3 with an accuracy of 0.280. Additionally, the quadratic weighted kappa (QWK) scores shown in Figure \ref{fig:radar-baw} indicate that LLama3.3 outperforms all other models in every category.

Comparing stability, LLama3.3 was able to provide a well-formed JSON response 101 out of 101 times. Mixtral failed to do so 44 times, DeepSeek 50 times, and Qwen 11 times. Due to these circumstances, as well as limited computing power, only LLama3.3 70b has been tested for other methods, and was used for all further results.

For the Most-similar-matches RAG approach, the QWK values indicate, that adding more context is not beneficial to the system. The single best match resulted in an accuracy of 0.308, the four best matches in 0.279, and seven best matches in 0.283.

\begin{figure}
    \centering
    \includegraphics[width=0.9\linewidth]{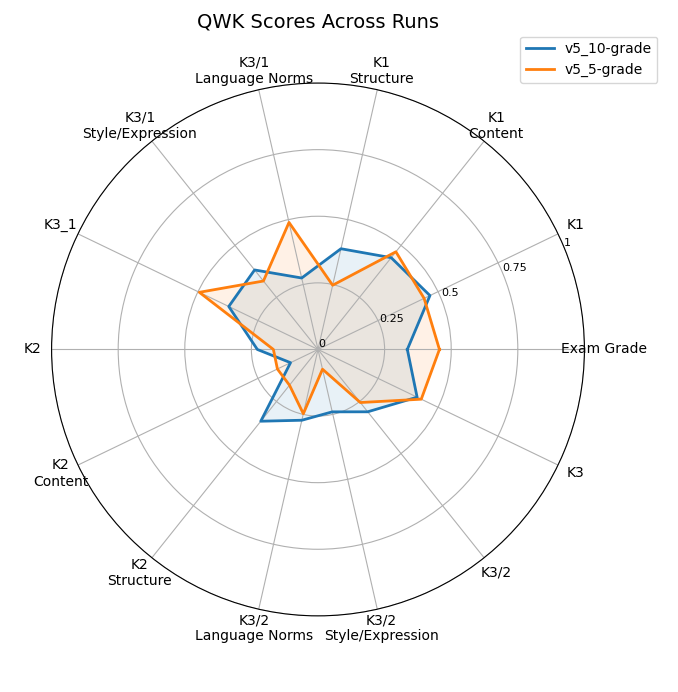}
    \caption{QWK accuracy for different RAG context sizes using sampling strategy "Range-of-examples"}
    \label{fig:radar-grade}
\end{figure}

For Range-of-examples, the differences of QWK score distribution in Figure \ref{fig:radar-grade} show a clear distinction between 5 and 10 texts. Adding 5 texts (orange) is beneficial for K1 related dimensions, while adding 10 texts (blue) increases performance mostly in K2 dimensions.

\begin{table}[htbp]
\centering
\caption{QWK Scores}
\label{tab:qwk-compressed}
\resizebox{\linewidth}{!}{
\begin{tabular}{lccc}
\toprule
Technique & Grade & Task 1 dimensions & Task 2 dimensions\\
\midrule
baseline       & .29 & .39/.09/.41/.19 & .07/.10/.23/.12 \\
RAG-1-best     & \textbf{.48} & .47/.22/.49/.39 & .19/.15/.30/.17 \\
RAG-10-grade   & .34 & .44/.39/.28/.38 & .12/\textbf{.34}/.27/.24 \\
RAG-4-best     & .47 & .45/.20/\textbf{.54}/.38 & .19/.20/.26/.17 \\
RAG-5-grade    & .46 & .47/.25/.49/.33 & .17/.17/.25/.08 \\
RAG-7-best     & .38 & .39/.21/.35/.29 & .19/.18/.37/.22 \\
RAG-best-worst & .46 & .41/.23/.50/.31 & .17/.17/.24/.13 \\
Few-all-grades & .25 & .31/.23/.25/.20 & .06/.06/.18/.16 \\
Few-best-worst & .40 & \textbf{.50}/.49/.36/.44 & \textbf{.22}/.28/.30/.25 \\
Few-mixed      & .43 & .49/\textbf{.55}/.50/\textbf{.50} & .20/.21/.28/\textbf{.30} \\
CoT-best-worst & .23 & .23/.30/.18/.34 & .04/.15/\textbf{.37}/.16 \\
\bottomrule
\end{tabular}}
{\footnotesize Dimensions: Content, Structure, Language Norms, Style/Expression}
\end{table}

\begin{table}[htbp]
\centering
\caption{Percentage Accuracy}
\label{tab:acc-compressed}
\resizebox{\linewidth}{!}{
\begin{tabular}{lccc}
\toprule
Technique & Grade & Task 1 & Task 2 \\
\midrule
baseline       & 20.8 & 24.8/27.7/27.7/28.7 & 21.8/22.8/24.8/21.8 \\
RAG-1-best     & 30.8 & 33.3/17.9/33.3/28.2 & \textbf{32.1}/21.8/23.1/28.2 \\
RAG-10-grade   & 30.2 & 30.2/30.2/29.2/31.2 & 25.0/\textbf{33.3}/22.9/33.3 \\
RAG-4-best     & 27.9 & 29.1/20.9/32.6/26.7 & 30.2/22.1/24.4/29.1 \\
RAG-5-grade    & 28.9 & 32.0/19.6/33.0/29.9 & 28.9/23.7/26.8/28.9 \\
RAG-7-best     & 28.3 & 28.3/29.3/22.2/28.3 & 26.3/27.3/25.3/27.3 \\
RAG-best-worst & 28.0 & 28.0/23.0/34.0/27.0 & 30.0/22.0/27.0/29.0 \\
Few-all-grades & 22.5 & 35.0/32.5/25.0/32.5 & 17.5/18.8/25.0/30.0 \\
Few-best-worst & 25.7 & \textbf{38.6}/39.6/\textbf{39.6}/32.7 & 20.8/20.8/18.8/24.8 \\
Few-mixed      & 24.8 & 36.6/\textbf{40.6}/33.7/38.6 & 26.7/28.7/22.8/30.7 \\
CoT-best-worst & \textbf{32.8} & 36.2/34.5/34.5/\textbf{39.7} & 29.3/29.3/\textbf{29.3}/\textbf{36.2} \\
\bottomrule
\end{tabular}}
{\footnotesize Dimensions: Content, Structure, Language Norms, Style/Expression}
\end{table}

\subsection{Technique comparison}
Across all four metrics, no clear best candidate or technique emerged. Few-shot prompting seems to yield the more promising results, despite the MAE and PCC score of RAG approaches outperforming those of few-shot. It needs to be taken into account, however, that the grading variety of few-shot prompts, and therefore the potential error, is larger than for RAG. 

\subsubsection{QWK}
Inspecting QWK scores, few-shot prompting overall, and specifically the method of mixing context lengths for short and longer text types, achieved the best results. Few-shot-mixed excelled in Task 1, with scores between 0.49 and 0.55, while Few-best-worst performed better in Task 2, with scores in the range of 0.22 to 0.30 for Task 2, and a QWK of 0.43 for the final grade.

The best QWK score for the final grade, however, was reached by RAG-1-best, where a single reference text was used for grading. Despite this, the other QWK scores for this technique can not compete with the best-in-class scores, casting doubt on the validity of this score.

CoT prompting with the Best-Average-Worst method could not achieve the standards of few-shot or even RAG without the additional reasoning effort. The only dimensions that worked especially well with CoT was language norms for shorter texts. As CoT adds more context due to, even internal, reasoning, it was expected that longer texts would benefit less from it, as the context window is already (over-)filled with reference and calibration texts.

\subsubsection{Accuracy}
The apparent failure of CoT when inspecting QWK scores is overshadowed by the percentage of absolute correct assigned grades, seen in Table \ref{tab:acc-compressed}. Even though the QWK score is suboptimal for CoT compared to its few-shot counterpart, the amount of correctly assigned grades is increased drastically for Task 2. As mentioned above, shorter texts seem to benefit from CoT approaches. For the final grade, CoT even reaches the highest agreement levels among model and human raters, significantly outperforming other few-shot approaches, as well as RAG implementations. 

RAG was not able to match the results achieved by few-shot prompting, despite the overall best QWK result for the final grade originating from a 1 context text RAG setup. When inspecting the individual dimensions scores of this setup, the accuracy of grader agreement is rather low, which gives cause to doubt the setup’s actual grading ability. As K3 is the average of the already averaged K3-sub-dimensions over both texts, numerical rounding and averaging might have helped this run. K3 is calculated the same way for all runs, however this one may have been particularly fortunate with how the calculations turned out. For validation, further runs are needed, which was not possible due to computational budget restrictions.  

\subsubsection{Mean Absolute Error (MAE) and Pearson Correlation Coefficient (PCC)}
Comparing scoring consistency using MAE, RAG tends to produce better results which stray less from the actual grade. Due to the aforementioned lack of grading variety, the better results of RAG are not as meaningful as it initially appears. 

An interesting division for PCC  can be observed in the sub-dimensions. For content, structure and language norms, RAG models are clearly the better choice, whilst for style and expression the performance of few-shot-mixed is best for long and short texts. QWK for K3/1 and K3/2 is also best rated by few-shot-mixed, hinting at a strong understanding of expression and style due to the iterative conversational build-up and alignment. For PCC, the same restrictions as for MAE apply. With larger errors possible due to a higher grading range, the potential for “extremely wrong” decisions is increased for few-shot approaches, potentially hindering the outcome.

\section{Discussion}
\label{section:Discussion}
The findings of this study provide important insights into the potential and current limitations of NLP-based automated essay scoring for Austrian A-levels. While significant progress was made through RAG and few-shot prompting, the results also highlight persistent challenges such as lack of variety in grading and computational restraints.

\subsection{LLM selection}
Initially this study contained four LLMs:
\begin{itemize}
    \item LLama3.3 70b
    \item DeepSeek-R1 32b
    \item Qwen3 30b
    \item Mixtral 8x7b
\end{itemize}

The models were chosen based on availability on Ollama \autocite{ollama2025website}, their supposed proficiency in the German language and the ability of the A100 GPU to support the models. 
During the course of evaluation, noticeable issues started to appear with all models but LLama3.3. Mixtral fell completely short of grasping the task, assigning the grade 3 to almost 100\% of candidate texts, rendering the model unusable for essay grading.

DeepSeek was able to perform more diverse grading, but was unable to adhere to the given reference texts. Its judgments were overly strict in grading, and did not match the feedback. Furthermore, on some occasions, the written feedback contained Chinese characters, notably only for negative feedback. 

In addition, both Mixtral and DeepSeek often failed to produce valid outputs, with failure rates upwards of 30\%. This unreliability disqualifies both models for usage in a real world environment.

Qwen was also very strict in its judgment, with the noticeable difference that bad marks were also followed up by detailed explanations that matched the given grade.

LLama3.3 had the best correlation and diversity in early experiments, with scores ranging from 2 to 4, but even this best model was unable to provide a grade of 1 without few-shot prompting.
LLama3.3's textual feedback is well-structured, and utilising few-shot calibration, the grading notes include detailed feedback for the candidate text.

A noticeable drawback for LLama3.3 is its computation time and resource intensity. Grading a pair of tasks with Qwen and Mixtral took about 30 seconds, DeepSeek doubled this time. LLama3.3, however, calculated about 240 seconds per task-pair, which increased up to 750 seconds for large context techniques.

Despite the drawback of time consumption, LLama3.3 had the overall best performance and was therefore deemed the LLM of choice for further investigation.

\subsection{Summary and Interpretation}
Before the conduct of the experiment, research on multi-text-type essay grading software for the German language was very limited. By assessing a total of 101 student exams for the Austrian A-levels in German, a generalisable, non-fine tuning set of methods has been proposed. Using the granular instructions of the SRDP committee and a standardised grading rubric, a set of seven types of texts can be graded by a single setup.

By providing LLama3.3 with information about the grading rubric, the assessment criteria and the importance of individual features for specific types of texts, a zero-shot baseline has been conducted, with results being expectedly suboptimal. The model mainly suffered from two problems, lack of grading variety and the absence of a sample grading process, which would allow them to reference a specific grade to a specific text with its strengths and weaknesses. 

The first solution uses RAG with three retrieval strategies: Best-Average-Worst, Most-similar-matches, and Range-of-examples. These improved the results over baseline, but lacked grading variety.

The second solution, few-shot prompting, outperformed RAG, allowing  LLama3.3 to grade on the full spectrum. Context length proved critical: Task 1, consisting of longer student texts, benefited from less context, while Task 2, made up of shorter texts, required more. A mixed, task-specific approach yielded the best results.

Adding CoT to few-shot prompting did not improve QWK, but raised percentage accuracy, especially on short texts. The lower QWK suggests mislabels were more severe, raising doubts whether higher agreement reflects real grading quality.

The authors also tried to add noisy context as proposed by \textcite{cuconasu2024power}. By inserting texts made up of random words the entropy and therefore grading quality should increase, which was not observable for this use case.

With the achieved results, fully automated grading is still a long way to go. The main issues for broad usage are the necessary computing power to self-host models, as well as the time it takes to receive an answer. For consumer applications, waiting upwards of 10 minutes for a currently mediocre grading result is not desirable for teachers and students alike. 

\subsection{Limitations}
\subsubsection{Dataset}
The dataset provided by the SRDP only contains three of the seven possible types of text, limiting the generalisation abilities of the conducted experiments. In addition, the process of removing personal data prevented the PDFs from being machine-readable and, in turn, introduced OCR-artifacts which, especially for RAG approaches, were treated as writing mistakes. Furthermore, all handwritten exams were not processable, introducing a further hurdle for an exam-wide rollout. 

A limitation that several studies in the area of AES face is the absence of multi-grader ground truths. For the SRDP dataset, only one human grader was responsible for the grades provided, potentially introducing subjective biases into the dataset. For future studies, regrading the dataset by at least one more human grader and observing potential differences would be a valuable addition, guaranteeing a more objective ground truth to test against.

\subsubsection{Models and computing power}
The A100 GPU available at the ASC was barely enough to run the largest model, LLama3.3 70b, in a reasonable amount of time. Evaluations for the complete dataset reached the 24-hour mark multiple times, which made iteration processes difficult and limited to a small dataset. These constraints potentially biased the development of strategies towards the preferences of this limited set of exams. Furthermore, the models in use are, except LLama3.3, not the largest available versions, therefore sacrificing potential.  

In general, most results per model have been achieved by a single test run, which, due to the inherent variance of LLMs, introduces an uncertainty into the results, as reproducibility is not guaranteed. Conducting more runs per method would produce a more grounded result, as well as providing information about the variance and stability of methods across multiple iterations.

These limitations do not undermine the value of the results but highlight the steps necessary for future studies: larger and more diverse datasets, multiple human graders, stronger compute resources, and validation with real teachers and students.

\section{Conclusion and Future Work}
\label{section:Conclusion}
The findings of this paper show that, while current LLMs cannot yet serve as fully autonomous graders for Austrian A-levels, they hold significant potential as supportive tools when combined with structured prompting, adaptive context strategies, and human oversight. It has been shown that even smaller models are capable of using pre-determined grading rubrics and standardised grading criteria in their evaluation processes, making the existence of a fully automated, human-level A-level grader only a matter of time.

To aid research in this direction, several points of work are proposed. For researchers with access to increased computing power, running larger models with multiple iterations per evaluation method would be a large step for German-based AES towards SOTA evaluation standards. To assess the reliability of the model's output, confidence scores can be implemented using majority voting or log-probs based approaches.

One major point for machine assisted or fully automated grading is the introduction of a fair and unbiased, standardised grading system, which does not favour a specific gender, social background, or other traits that should not have an influence on an exam's mark. The existence of disabilities, for which special grading criteria exist, must, however, be taken into consideration. It is also important to assess the model's robustness against malicious actors, as future students could hide prompt injections or similar attacks in their essay to earn an undeserved good grade. 

In conclusion, these findings underline that the future of AES lies not in replacing human judgment, but in creating reliable, fair, and transparent AI systems that enhance the work of educators and support student learning.

\begin{acks}
The work is supported by the Austrian Research Promotion Agency (FFG) through the project FAIR-AI, Grant no. 904624 \url{https://projekte.ffg.at/projekt/4847537}
\end{acks}

\printbibliography

@article{deerwester1990indexing,
  title={Indexing by latent semantic analysis},
  author={Deerwester, Scott and Dumais, Susan T and Furnas, George W and Landauer, Thomas K and Harshman, Richard},
  journal={Journal of the American society for information science},
  volume={41},
  number={6},
  pages={391--407},
  year={1990},
  publisher={Wiley Online Library}
}

@article{pack2024large,
  title={Large language models and automated essay scoring of English language learner writing: Insights into validity and reliability},
  author={Pack, Austin and Barrett, Alex and Escalante, Juan},
  journal={Computers and Education: Artificial Intelligence},
  volume={6},
  pages={100234},
  year={2024},
  publisher={Elsevier}
}

@misc{bmbwf2020beurteilung,
  author       = {{Bundesministerium für Bildung, Wissenschaft und Forschung}},
  title        = {Textsortenkatalog zur SRDP in der Unterrichtssprache (Deutsch,
Kroatisch, Slowenisch, Ungarisch)},
  year         = {2020},
  url          = {https://www.matura.gv.at/index.php?eID=dumpFile&t=f&f=4525&token=950c7f2b86f0ebc3459c5f0aa0e04013ab99c572},
  lastaccessed = {August 27, 2025}        
}

@misc{bmbwf2024angaben,
  author       = {{Bundesministerium für Bildung, Wissenschaft und Forschung}},
  title        = {Standardisierte kompetenzorientierte schriftliche
Reifeprüfung / Reife- und Diplomprüfung / Berufsreifeprüfung - 2. Mai 2024},
  year         = {2024},
  url          = {https://www.matura.gv.at/index.php?eID=dumpFile&t=f&f=6941&token=42e50976a60ff1c075d870c400d6794f161edd7c},
  lastaccessed = {August 27, 2025}  
}

@misc{bmbwf2023angaben,
  author       = {{Bundesministerium für Bildung, Wissenschaft und Forschung}},
 title        = {Standardisierte kompetenzorientierte schriftliche
Reifeprüfung / Reife- und Diplomprüfung / Berufsreifeprüfung - 5. Mai 2023},
  year         = {2023},
  url          = {https://www.matura.gv.at/index.php?eID=dumpFile&t=f&f=5836&token=cc7ef226aa5da503dd0d7966e4dc901cb6ba9c35},
  lastaccessed = {August 27, 2025}
}

@online{ollama2025website,
  title        = {Ollama: Run large language models locally},
  author       = {{Ollama}},
  year         = {2025},
  url          = {https://ollama.com/},
  lastaccessed = {August 27, 2025}
}

@inproceedings{cuconasu2024power,
  title={The power of noise: Redefining retrieval for rag systems},
  author={Cuconasu, Florin and Trappolini, Giovanni and Siciliano, Federico and Filice, Simone and Campagnano, Cesare and Maarek, Yoelle and Tonellotto, Nicola and Silvestri, Fabrizio},
  booktitle={Proceedings of the 47th International ACM SIGIR Conference on Research and Development in Information Retrieval},
  pages={719--729},
  year={2024}
}

@article{brown2020language,
  title={Language models are few-shot learners},
  author={Brown, Tom and Mann, Benjamin and Ryder, Nick and Subbiah, Melanie and Kaplan, Jared D and Dhariwal, Prafulla and Neelakantan, Arvind and Shyam, Pranav and Sastry, Girish and Askell, Amanda and others},
  journal={Advances in neural information processing systems},
  volume={33},
  pages={1877--1901},
  year={2020}
}

@article{lee2024applying,
  title={Applying large language models and chain-of-thought for automatic scoring},
  author={Lee, Gyeong-Geon and Latif, Ehsan and Wu, Xuansheng and Liu, Ninghao and Zhai, Xiaoming},
  journal={Computers and Education: Artificial Intelligence},
  volume={6},
  pages={100213},
  year={2024},
  publisher={Elsevier}
}

@article{page1966imminence,
  title={The imminence of... grading essays by computer},
  author={Page, Ellis B},
  journal={The Phi Delta Kappan},
  volume={47},
  number={5},
  pages={238--243},
  year={1966},
  publisher={JSTOR}
}

@preprint{xiao2024automation,
  title={From Automation to Augmentation: Large Language Models Elevating Essay Scoring Landscape},
  author={Xiao, Changrong and Ma, Wenxing and Xu, Sean Xin and Zhang, Kunpeng and Wang, Yufang and Fu, Qi},
  journal={arXiv preprint arXiv:2401.06431},
  year={2024}
}

@article{mizumoto2023exploring,
  title={Exploring the potential of using an AI language model for automated essay scoring},
  author={Mizumoto, Atsushi and Eguchi, Masaki},
  journal={Research Methods in Applied Linguistics},
  volume={2},
  number={2},
  pages={100050},
  year={2023},
  publisher={Elsevier}
}

@inproceedings{schaller2024fairness,
  title={Fairness in automated essay scoring: A comparative analysis of algorithms on German learner essays from secondary education},
  author={Schaller, Nils-Jonathan and Ding, Yuning and Horbach, Andrea and Meyer, Jennifer and Jansen, Thorben},
  booktitle={Proceedings of the 19th workshop on innovative use of nlp for building educational applications (bea 2024)},
  pages={210--221},
  year={2024}
}

@inproceedings{sessler2025can,
  title={Can AI grade your essays? A comparative analysis of large language models and teacher ratings in multidimensional essay scoring},
  author={Se{\ss}ler, Kathrin and F{\"u}rstenberg, Maurice and B{\"u}hler, Babette and Kasneci, Enkelejda},
  booktitle={Proceedings of the 15th International Learning Analytics and Knowledge Conference},
  pages={462--472},
  year={2025}
}

@article{mussmann2016niedersachsische,
  title={Nieders{\"a}chsische Arbeitszeitstudie Lehrkr{\"a}fte an {\"o}ffentlichen Schulen 2015/2016},
  author={Mu{\ss}mann, Frank and Riethm{\"u}ller, Martin and Hardwig, Thomas and Ohms, Ilka Charlotte and Kl{\"o}tzer, Stefan},
  journal={Unter Mitarbeit von Stefan Peters, Marcel Parciak, Ilka Charlotte Ohms und Stefan Kl{\"o}tzer. G{\"o}ttingen: Georg-August-Universit{\"a}t G{\"o}ttingen, Kooperationsstelle Hochschulen und Gewerkschaften},
  year={2016}
}

@misc{hamner2012asap,
  author       = {Ben Hamner and Jaison Morgan and Iynnvandev and Mark Shermis and Tom Vander Ark},
  title        = {The Hewlett Foundation: Automated Essay Scoring},
  year         = {2012},
  url = {https://www.kaggle.com/c/asap-aes},
  lastaccessed = {August 27, 2025},
}

@preprint{stahl2024exploring,
  title={Exploring LLM prompting strategies for joint essay scoring and feedback generation},
  author={Stahl, Maja and Biermann, Leon and Nehring, Andreas and Wachsmuth, Henning},
  journal={arXiv preprint arXiv:2404.15845},
  year={2024}
}

@preprint{mikolov2013efficient,
  title={Efficient estimation of word representations in vector space},
  author={Mikolov, Tomas and Chen, Kai and Corrado, Greg and Dean, Jeffrey},
  journal={arXiv preprint arXiv:1301.3781},
  year={2013}
}

@article{vaswani2017attention,
  title={Attention is all you need},
  author={Vaswani, Ashish and Shazeer, Noam and Parmar, Niki and Uszkoreit, Jakob and Jones, Llion and Gomez, Aidan N and Kaiser, {\L}ukasz and Polosukhin, Illia},
  journal={Advances in neural information processing systems},
  volume={30},
  year={2017}
}

@article{burstein2003rater,
  title={The e-rater scoring engine: Automated essay scoring with natural language processing},
  author={Burstein, Jill},
  journal={Automated essay scoring: A cross-disciplinary perspective},
  volume={113121},
  year={2003}
}

@article{elliot2003intellimetric,
  title={IntelliMetric: From here to validity},
  author={Elliot, Scott},
  journal={Automated essay scoring: A cross-disciplinary perspective},
  pages={71--86},
  year={2003},
  publisher={Hillsdale}
}

@inproceedings{dong2016automatic,
  title={Automatic features for essay scoring--an empirical study},
  author={Dong, Fei and Zhang, Yue},
  booktitle={Proceedings of the 2016 conference on empirical methods in natural language processing},
  pages={1072--1077},
  year={2016}
}

@inproceedings{taghipour2016neural,
  title={A neural approach to automated essay scoring},
  author={Taghipour, Kaveh and Ng, Hwee Tou},
  booktitle={Proceedings of the 2016 conference on empirical methods in natural language processing},
  pages={1882--1891},
  year={2016}
}

@inproceedings{dong2017attention,
  title={Attention-based recurrent convolutional neural network for automatic essay scoring},
  author={Dong, Fei and Zhang, Yue and Yang, Jie},
  booktitle={Proceedings of the 21st conference on computational natural language learning (CoNLL 2017)},
  pages={153--162},
  year={2017}
}

@inproceedings{devlin2019bert,
  title={Bert: Pre-training of deep bidirectional transformers for language understanding},
  author={Devlin, Jacob and Chang, Ming-Wei and Lee, Kenton and Toutanova, Kristina},
  booktitle={Proceedings of the 2019 conference of the North American chapter of the association for computational linguistics: human language technologies, volume 1 (long and short papers)},
  pages={4171--4186},
  year={2019}
}

@article{firoozi2025using,
  title={Using automated procedures to score educational essays written in three languages},
  author={Firoozi, Tahereh and Mohammadi, Hamid and Gierl, Mark J},
  journal={Journal of Educational Measurement},
  volume={62},
  number={1},
  pages={33--56},
  year={2025},
  publisher={Wiley Online Library}
}

@preprint{feng2020language,
  title={Language-agnostic BERT sentence embedding},
  author={Feng, Fangxiaoyu and Yang, Yinfei and Cer, Daniel and Arivazhagan, Naveen and Wang, Wei},
  journal={arXiv preprint arXiv:2007.01852},
  year={2020}
}

@article{ludwig2021automated,
  title={Automated essay scoring using transformer models},
  author={Ludwig, Sabrina and Mayer, Christian and Hansen, Christopher and Eilers, Kerstin and Brandt, Steffen},
  journal={Psych},
  volume={3},
  number={4},
  pages={897--915},
  year={2021},
  publisher={MDPI}
}

@preprint{stahl2024school,
  title={A school student essay corpus for analyzing interactions of argumentative structure and quality},
  author={Stahl, Maja and Michel, Nadine and Kilsbach, Sebastian and Schmidtke, Julian and Rezat, Sara and Wachsmuth, Henning},
  journal={arXiv preprint arXiv:2404.02529},
  year={2024}
}

@preprint{he2021debertav3,
  title={Debertav3: Improving deberta using electra-style pre-training with gradient-disentangled embedding sharing},
  author={He, Pengcheng and Gao, Jianfeng and Chen, Weizhu},
  journal={arXiv preprint arXiv:2111.09543},
  year={2021}
}

@preprint{touvron2023llama,
  title={Llama: Open and efficient foundation language models},
  author={Touvron, Hugo and Lavril, Thibaut and Izacard, Gautier and Martinet, Xavier and Lachaux, Marie-Anne and Lacroix, Timoth{\'e}e and Rozi{\`e}re, Baptiste and Goyal, Naman and Hambro, Eric and Azhar, Faisal and others},
  journal={arXiv preprint arXiv:2302.13971},
  year={2023}
}

\end{document}